\documentclass[runningheads]{llncs}
\usepackage{comment}
\usepackage{amssymb, amsmath}

\usepackage[T1]{fontenc}
\usepackage{graphicx}
\newcommand{\A}{\mathcal{A}}
\newcommand{\C}{\mathbb{C}}
\newcommand{\F}{\mathcal{F}}

\begin{document}
\title{Universal kernels via harmonic analysis on Riemannian symmetric spaces}
%
%

\author{Franziskus Steinert\inst{1} \and
Salem Said\inst{2}
\and
Cyrus Mostajeran\inst{1}\orcidID{0000-0001-8910-9755}}
\authorrunning{F. Steinert et al.}
%
\institute{School of Physical and Mathematical Sciences, Nanyang Technological University, 21 Nanyang Link, Singapore 637371, Singapore \\
\email{cyrussam.mostajeran@ntu.edu.sg}
\and
Laboratoire Jean Kuntzman, Université Grenoble-Alpes, Grenoble 38400, France}

\maketitle              

\begin{abstract}
The universality properties of kernels characterize the class of functions that can be approximated in the associated reproducing kernel Hilbert space and are of fundamental importance in the theoretical underpinning of kernel methods in machine learning. In this work, we establish fundamental tools for investigating universality properties of kernels in Riemannian symmetric spaces, thereby extending the study of this important topic to kernels in non-Euclidean domains. Moreover, we use the developed tools to prove the universality of several recent examples from the literature on positive definite kernels defined on Riemannian symmetric spaces, thus providing theoretical justification for their use in applications involving manifold-valued data.

\end{abstract}

\section{Introduction}
Kernel methods are widely used in machine learning for their ability to model nonlinear structures via embeddings in reproducing kernel Hilbert spaces (RKHS). A key property of kernels is universality which characterizes the algorithm's capability to learn arbitrarily complex patterns. More precisely, a kernel $k$ on a space $X$ is said to be universal if its RKHS, i.e. the closure of
\begin{equation} \label{RKHS}
    \mathcal{K}:= \bigg\{ \sum_{j=1}^N   a_j\,k(x_j,\cdot): a_j\in \C, x_j\in X, N\geq 1\bigg\}
\end{equation}
under the inner product
\begin{equation} \label{innerprodrkhs}
    \bigg\langle \sum_{i=1}^Na_i \, k(x_i,\cdot),\, \sum_{j=1}^M b_j \, k(y_j,\cdot) \bigg\rangle_\mathcal{K} := \sum_{i,j} a_i\,  b_j^* \, k(x_i, y_j),
\end{equation}
is dense in a certain function space with respect to a given topology. Here, $^*$ denotes complex conjugation.

While universality is well established for Euclidean spaces, its extension to non-Euclidean settings is less developed. However, in recent years, manifold-valued data has become increasingly important in fields such as computer vision \cite{huang2017riemannian,lopez2021vector}, medicine \cite{arsigny2007geometric,liu2019hyperbolic} or natural language processing \cite{chami2019hyperbolicgraphconvolutionalneural}. Inferior performance of unadapted Euclidean algorithms compared to their geometry-aware generalizations on such data \cite{pennec2006riemannian} further increased the need for incorporating geometric considerations into algorithm design.

An important challenge in extending kernel methods to non-Euclidean domains has been to ensure that the kernels are positive definite, which is critical for the existence of an RKHS. This is further highlighted by the observation that the obvious generalization of the so-called Gaussian or radial basis function kernel to Riemannian manifolds is often not positive definite \cite{jayasumana2015kernel,Feragen_2015_CVPR,azangulov2024stationary,dacosta2024invariantkernelsriemanniansymmetric,Dacosta2024JMLR}. Recent works have offered solutions to this challenge and provided systematic techniques for generating positive definite kernels on Riemannian manifolds \cite{Boroviskiy2020,azangulov2024stationary,dacosta2024invariantkernelsriemanniansymmetric,mostajeran2024invariantkernelsspacecomplex}.

Here, we complement the recent literature on the topic of kernels on manifolds by establishing fundamental results concerning the universality of kernels on Riemannian symmetric spaces. In particular, we develop the tools necessary to establish the universality of the heat and Matérn kernels on symmetric spaces in addition to the kernels introduced in \cite{dacosta2024invariantkernelsriemanniansymmetric,mostajeran2024invariantkernelsspacecomplex} using tools from harmonic analysis. The latter are particularly attractive as they offer more computationally tractable closed-form alternatives to the heat and Matérn kernels while maintaining the key properties of positive definiteness and universality as we shall see.

To keep the presentation brief, we will focus on the case of symmetric spaces of non-compact type.



\section{Background and Notation} \label{background}

\subsection{Kernels via spectral densities}

A Riemannian symmetric space is a Riemannian manifold $X$ that admits a geodesic-reversing isometry about each one of its points~\cite{helgasonsymmetric}. 
If $G$ is the identity component of the isometry group of $X$ and $K \subset G$ the stabiliser of some point $o \in X$, then $G$ acts transitively on $X$ and one may identify $X$ with the coset space $G/K$. In addition, the Lie algebra $\mathfrak{g}$ of $G$ admits a Cartan decomposition $\mathfrak{g} = \mathfrak{k} + \mathfrak{p}$ where $\mathfrak{k}$ is the Lie algebra of $K$ and $\mathfrak{p}$ is a subspace of $\mathfrak{g}$, which is isomorphic to the tangent space $T_o\hspace{0.02cm}X$. Here, it will be assumed that $X$ is a symmetric space of non-compact type\,: $\mathfrak{g}$ is a real semisimple Lie algebra whose Killing form is 
negative-definite on $\mathfrak{k}$ and positive-definite on $\mathfrak{p}$. Geometrically, $X$ is then diffeomorphic to $\mathbb{R}^n$ ($n = \dim X$) and negatively curved~\cite{helgasonsymmetric}.

Let $\mathfrak{a}$ be a maximal Abelian subspace of $\mathfrak{p}$ and $\Sigma$ the set of restricted roots of $\mathfrak{g}$ with respect to $\mathfrak{a}$. Choose a set $\Sigma_+$ of positive roots and a corresponding Weyl chamber $\mathfrak{a}_+$. Recall that each $g \in G$ admits a unique Iwasawa decomposition~\cite{helgasonsymmetric} 
\begin{equation*}
    g= n(g)\, \exp(a(g))\, k(g) \quad \text{where } n(g)\in N, a(g)\in \mathfrak{a}, k(g)\in K
\end{equation*}
and $N$ is the Lie subgroup of $G$ corresponding to $\mathfrak{n}$ (the nilpotent subalgebra of $\mathfrak{g}$ equal to the sum of root subspaces $\mathfrak{g}_{\hspace{0.02cm}\sigma}$ where $\sigma \in \Sigma_+$). Further, let $M$ be the centraliser of $\mathfrak{a}$ in $K$ and $B = K/M$. Denote by $d\nu(\lambda,b) = |c(\lambda)|^{-2}\, d\lambda\,db$ the Plancherel measure on $\A:= \mathfrak{a}^\ast_+ \times B$, where $\mathfrak{a}^\ast$ is the dual of $\mathfrak{a}$ (note~that $\mathfrak{a}^\ast$ is identified with $\mathfrak{a}$ through the Killing form). Here, $c(\lambda)$ denotes the Harish-Chandra function~\cite{helgasonthirdbook}. 

Finally, let $e(g,\lambda,b) = e^{(i\lambda -\rho)H(g^{-1}b)}$ for $g \in G$ and $(\lambda,B) \in \mathcal{A}$, where $H(g)= -a(g^{-1})$ and $\rho$ is half the sum of the positive roots $\sigma \in \Sigma_+$ (counted with multiplicities).
This only depends on $x = g\cdot o$ (the dot denotes the action of $g \in G$ on $o \in X$), and may then be written $e(x,\lambda,b)$. 

With this notation, the general form of the kernels $k$ considered in the present work is given by the following definition.

\begin{definition} \label{definitionkernel}
Let $\psi \in L^1 (\A,\nu) \cap L^2(\A,\nu)$ and define the map $k:X\times X \rightarrow \mathbb{C}$
\begin{equation} \label{kernelnocomp}
        k(x,y) = \int_\A \psi(\lambda, b) \, e(x,\lambda,b) \, e^*(y,\lambda,b) \,  d\nu(\lambda,b)\\
\end{equation}
where the $^*$ denotes complex conjugation.
\end{definition}

The significance of this formulation is that the properties of a kernel $k$ given by (\ref{kernelnocomp}) are determined by its spectral density $\psi$. A first result in this direction is the following, where we recall that a kernel $k$ is said to be $G$-invariant if $k(g\cdot x, \, g\cdot y)= k(x,y)$ for all $g\in G$ and $x,y\in X$.  


\begin{lemma} \label{kernelprops}
    If $\psi \geq 0$ then the kernel $k$ in (\ref{kernelnocomp}) is Hermitian and positive-definite. Moreover, if $\psi(\lambda,b)$ is independent of $b \in B$ then $k$ is $G$-invariant. 
\end{lemma} 

\subsection{The Helgason-Fourier transform}

To go beyond these general properties and establish universality properties, it will be helpful to introduce the tools of harmonic analysis on $X$, especially the Helgason-Fourier transform~\cite{helgasonthirdbook}. 

The Helgason-Fourier transform of a compactly supported continuous function $f:X \rightarrow \mathbb{C}$ is a continuous function $\F [f]:\A \rightarrow \mathbb{C}$~\cite{helgasonthirdbook} (Chapter III),
\begin{equation*}
     \F [f](\lambda,b):= \int_X f(x)\, e(x,\lambda,b) \, \mathrm{vol}(dx)
\end{equation*}
where $\mathrm{vol}(dx)$ denotes the Riemannian volume measure on $X$. The Helgason-Fourier transform extends to an isomorphism of Hilbert spaces $\F: L^2(X,\mathrm{vol}) \to L^2(\A,\nu)$ \cite{helgasonthirdbook}. If the function $f$ is $K$-invariant, the Helgason-Fourier transform reduces to the spherical transform $\hat{f}$. This property bridges the gap between the generalized treatment using the Helgason-Fourier transform in the present paper and the approach based on spherical transforms in \cite{dacosta2024invariantkernelsriemanniansymmetric}. In particular, the $L^2$-Godement theorem of \cite{dacosta2024invariantkernelsriemanniansymmetric} can be recast as a partial converse of Lemma \ref{kernelprops}. 
\begin{proposition} \label{l2godement}
   Any square-integrable $G$-invariant 
   positive-definite kernel $k$ is given by (\ref{kernelnocomp}), with $\psi \geq 0$ and $\psi(\lambda,b)$ independent of $b \in B$. 
\end{proposition}
Here, a kernel $k:X \times X \rightarrow \mathbb{C}$ is called square-integrable if $f(x) = k(o,x)$ is in $L^2(X,\mathrm{vol})$. Proposition \ref{l2godement} shows that kernels of the form (\ref{kernelnocomp}) are quite general, and therefore justifies the choice of (\ref{kernelnocomp}) as a starting point for further study.

\subsection{Sobolev space theory}
This will be required in order to obtain universality properties for continuous functions. Sobolev spaces are here defined in accordance with \cite{kassymov2024functionalinequalitiessymmetricspaces}. 

Let $X$ be a complete Riemannian manifold, $\Delta$ its Laplace-Beltrami operator. For $s > 0$, let
\begin{equation*} \label{definitionsobolev}
    H^{s}(X):= \bigg\{ u\in L^2(X,\mathrm{vol}): \, (\Delta)^{\frac{s}{2}} u\in L^2(X,\mathrm{vol}) \bigg\}
\end{equation*}
be endowed with the norm $\| u\|_{H^{s}(X)} := \, \|(\Delta)^{\frac{s}{2}}u\|_{L^2(X)} + \| u\|_{L^2(X)}$. One has the following version of the
Sobolev embedding theorem~\cite{krupka2011handbook}. \\
\begin{theorem} \label{SET}
Assume $X$ has Ricci curvature bounded below and positive injectivity radius. If $s>m+\frac{n}{2}$, then there exists a continuous embedding $H^{s}(X) \hookrightarrow \mathcal{C}^m_b(X)$. \\
\end{theorem}
Here, $\mathcal{C}^m_b(X)$ is the space of bounded $C^m$ functions on $X$ with the norm defined in~\cite{krupka2011handbook}.
Roughly, this is the sum of supremum norms of all derivatives of order up to $m$.

If $X$ is a symmetric space of non-compact type, $X$ satisfies the assumptions of Theorem \ref{SET}, because $X$ is always complete and homogeneous. Moreover, there is a spectral characterisation
(see~\cite{pesenson2008discrete}), 
\begin{equation} \label{eq:sobol_spectral}
    f\in H^s(X) \iff \langle \lambda \rangle^s \, \F [f] \in L^2(\mathcal{A},\nu)
\end{equation}
where $\langle \lambda \rangle = (\| \lambda\|^2 + \| \rho\|^2)^{1/2}$ with the norms $\|\cdot\|$ induced by the Killing form. Note that $-\langle \lambda \rangle^2$ are the eigenvalues of $\Delta$ corresponding to the eigenfunctions $e(x,\lambda,b)$~\cite{helgason2022groups}.

\section{Main results} \label{main}

\subsection{$L^2$-universality} \label{l2case}

The cornerstone of our approach is to establish an alternative form of the RKHS (\ref{RKHS}) of $k$ that describes the RKHS in the Fourier domain rather than in terms of translates of the kernel.

\begin{definition} \label{defl2univ}
    A kernel $k: X\times X \to \C$ is said to be $L^2$-universal, if its RKHS (\ref{RKHS}) is dense in $L^2(X,\mathrm{vol})$ with respect to the $L^2$-norm.
\end{definition}
Suppose now that the spectral density $\psi$ in (\ref{kernelnocomp}) is in fact strictly positive $\nu$-a.e. and define the space
\begin{equation} \label{rkhsnc}
    \mathcal{H}_k := \left\{ f\in L^2(X): \frac{\F[f]}{\sqrt{\psi}} \in L^2(\mathcal{A}) \right\}.
\end{equation}
The following lemma states that $\mathcal{H}_k$ is an RKHS with reproducing kernel (\ref{kernelnocomp}).Then, by the one-to-one correspondence between kernels and their reproducing kernel Hilbert spaces \cite{aronszajnrkhs}, the~space $\mathcal{H}_k$ is \textit{the} RKHS of $k$.

In the following, $L^2(X,\mathrm{vol})$ and $L^2(\mathcal{A},\nu)$ will be denoted $L^2(X)$ and $L^2(\mathcal{A})$, respectively. 
\begin{lemma} \label{lemmarkhs}
    Let $\psi\in L^2(\A)$ be bounded, $\nu$-integrable, strictly positive $\nu$-a.e., and set $\eta := \sqrt{\psi}$. For the space $\mathcal{H}_k$ given by (\ref{rkhsnc}), define the map $T: \mathcal{H}_k \to L^2(X)$ by $$f\mapsto \F^{-1}[\F[f]/\eta].$$ Then, $\mathcal{H}_k$~is a Hilbert space with inner product $$\langle f,g\rangle_k:= \langle Tf, Tg\rangle_{L^2(X)}$$ and reproducing kernel (\ref{kernelnocomp}). \\
\end{lemma}

Note that any $h\in \mathcal{C}_c(\A)$ satisfies the integrability requirement of $\mathcal{H}_k$. Hence, any inverse transform of a compactly supported continuous function is contained in the RKHS. However, compactly supported continuous functions are dense in $L^2(\A)$, so~that the Plancherel theorem yields the following result:

\begin{theorem} \label{l2universality}
    If $\psi \in L^2(\A)$ is independent of $b\in B$ and satisfies the assumptions of Lemma \ref{lemmarkhs}, then the kernel (\ref{kernelnocomp}) is $L^2$-universal.
\end{theorem}

Note that according to Theorem~\ref{l2universality}, functions that are not $K$-invariant can still be learned through a $K$-invariant kernel. This phenomenon occurs since translates of a $K$-invariant kernel are not $K$-invariant anymore.

\subsection{Universality for continuous functions}
Building on the $L^2$-case, we employ Sobolev space theory to strengthen the universality properties of the kernel (\ref{kernelnocomp}). There are several notions of universality for continuous functions, depending on the underlying topology. The first one we consider is the following:

\begin{definition} \label{defccuniv}
    A kernel $k: X\times X \to \C$ is said to be $\mathcal{C}_c$-universal, if it is continuous and its RKHS (\ref{RKHS}) is dense in $\mathcal{C}(X)$ with respect to the topology of local uniform convergence.
\end{definition}

\noindent Using the description of the RKHS afforded by Lemma \ref{lemmarkhs} together with Sobolev embeddings \cite{krupka2011handbook}, kernels whose spectral density $\psi$ satisfies certain decay conditions can be shown to be $\mathcal{C}_c$-universal. This result is captured in Theorem \ref{ccuniversality} below where 
$\langle \lambda \rangle = (\| \lambda\|^2 + \| \rho\|^2)^{1/2}$ with the norms induced by the Killing form. Here, $-\langle \lambda \rangle^2$ are the eigenvalues of the Laplace-Beltrami operator on $X$ corresponding to the eigenfunctions $e(x,\lambda,b)$.

\begin{theorem} \label{ccuniversality}
    Suppose that, in addition to the requirements in Theorem \ref{l2universality}, $\psi$ satisfies the decay condition $\psi(\lambda) = \mathcal{O}(\langle \lambda \rangle^{-2s})$ for some $s> n/2$ ($n = \dim X$). Then, the kernel (\ref{kernelnocomp}) is $\mathcal{C}_c$-universal. 
\end{theorem}

\noindent However, a drawback of this notion of universality is its locality~\cite{sriperumbudur2011universality}. The following definition provides a partial remedy by using a stronger topology albeit considering a smaller target function space.

\begin{definition} \label{defc0univ}
\begin{enumerate}
    \item A function $f: X\to \C$ is said to belong to $\mathcal{C}_0(X)$ if for any $\epsilon >0$ the set $\{ x\in X: |f(x)|\geq \epsilon \}$ is compact.
    \item A kernel $k:X\times X\to \C$ is said to be $\mathcal{C}_0$-universal, if for any $x\in X$ one has $k(x,\cdot) \in \mathcal{C}_0(X)$ and its RKHS (\ref{RKHS}) is dense in $\mathcal{C}_0(X)$ with respect to the topology of uniform convergence.
\end{enumerate}
\end{definition}

\noindent Using almost the same arguments as for Theorem \ref{ccuniversality}, one may prove the following:

\begin{corollary} \label{c0universality}
    Suppose that, in addition to the requirements in Theorem \ref{ccuniversality}, the kernel $k$ satisfies $k(x,\cdot) \in \mathcal{C}_0(X)$ for any $x\in X$. Then, $k$ is $\mathcal{C}_0$-universal.
\end{corollary}

\section{Examples} \label{examples}

Here we investigate the universality of two important classes of kernels on symmetric spaces by identifying the associated spectral densities and using
Theorems \ref{l2universality} and \ref{ccuniversality}.

\subsection{Heat and Matérn kernels} \label{subsec:heatmatern}
The heat and Matérn kernels are natural kernels on any symmetric space $X$, which arise from the study of the heat equation on $X$. In the notation of (\ref{kernelnocomp}), they are given by the spectral densities~\cite{azangulov2024stationary}
\begin{align}
\label{heatdensity}    \psi(\lambda) & = c_\kappa\, \exp(-\frac{\kappa^2}{2}\langle \lambda \rangle^2) & \text{(heat kernel)}\\
\label{materndensity}    \psi(\lambda) & = c_{\kappa, \nu} \left( \frac{2\nu}{\kappa^2}  +\langle \lambda \rangle^2 \right)^{-\nu -n/2} & \text{(Matérn kernel)}
\end{align}
where $c_\kappa, c_{\nu, \kappa}>0$ are constants of normalisation, while $\kappa$ and $\nu$ are known as the length-scale and smoothness parameters, respectively. Theorems \ref{l2universality} and \ref{ccuniversality}, along with Corollary \ref{c0universality}, make it clear both of these kernels are $L^2$-universal and $\mathcal{C}_c$-universal. However, a significant drawback remains, since the evaluation of these kernels is computationally expensive and sometimes numerically unstable.

\subsection{Beta-prime kernels} \label{subsec:betaprime}
Besides the heat and Matérn kernels, the second class of kernels considered here is that of Beta-prime kernels. These can be defined when $X$ is a symmetric cone, such as the cone of real, complex, or quaternion covariance matrices~\cite{farautcones} (\textit{i.e.} $G = GL(n,\mathbb{K})$ and $K = U(n,\mathbb{K})$ with $\mathbb{K} = \mathbb{R}$, $\mathbb{C}$, or $\mathbb{H}$, and $U(n,\mathbb{K})$ the orthogonal, unitary or symplectic group, respectively).

Let $X$ be any one of these three spaces (cones) of covariance matrices. The Beta-prime kernel is given directly by its analytic expression
\begin{equation} \label{eq:betaprime}
  k(x,y) = \left(\frac{\det(x)\det(y)}{\det(x+y)^2}\right)^{\!\alpha}
\end{equation}
where $\alpha > \frac{\beta}{2}(n-1)$ ($\beta = 1,2, 4$ according to $\mathbb{K} = \mathbb{R},\mathbb{C},\mathbb{H}$) and
the determinant of a quaternion matrix is defined according to~\cite{quaterniondet}. It can be shown that 
$f(x) = k(o,x)$ is in $L^1(X,\mathrm{vol})\cap L^2(X,\mathrm{vol})$ and that $k$ is a kernel with spectral density
\begin{equation} \label{eq:psibetaprime}
  \psi(\lambda) = \frac{1}{\Gamma_X(2\alpha)}
  \left|\Gamma_X(\alpha +\rho + \mathrm{i}\lambda)\right|^2
\end{equation}
where $\Gamma_X:\mathfrak{a}^\ast_{\mathbb{C}} \rightarrow \mathbb{C}$ is the so-called cone Gamma function~\cite{farautcones} ($\mathfrak{a}^\ast_{\mathbb{C}}$ denotes the complexification of $\mathfrak{a}^\ast$). Now, $\mathfrak{a}^\ast_{\mathbb{C}}$ can be identified in a natural way with $\mathbb{C}^n$, so that
$$   
\Gamma_X(\tau_1,\ldots,\tau_n) = (2\pi)^{(d-n)/2}\prod^n_{j=1}\Gamma\left(\tau_j - \frac{\beta}{2}(j-1)\right)
$$
for $(\tau_1,\ldots,\tau_n) \in \mathbb{C}^n$, where $d$ is the dimension of $X$. Here, $\Gamma:\mathbb{C} \rightarrow \mathbb{C} \cup \lbrace \infty \rbrace$ denotes the Euler Gamma function. 

Given the spectral density (\ref{eq:psibetaprime}), it is straightforward to apply Theorems \ref{l2universality} and \ref{ccuniversality}. First, since the Euler Gamma function admits no zeros in the complex plane, Theorem \ref{l2universality} shows that Beta-prime kernels are $L^2$-universal. Second, recalling the classical asymptotic expansion~\cite{wongbeals}
$$
\left|\Gamma(x+\mathrm{i}y)\right| = \sqrt{2\pi}\hspace{0.02cm}y^{x-\frac{1}{2}}\hspace{0.02cm}e^{-\frac{\pi}{2}|y|}\left(1+O(|y|^{-1}) \right)\hspace{1cm}
x,y \in \mathbb{R} $$
it becomes clear the decay condition in Theorem \ref{ccuniversality} holds for the spectral density (\ref{eq:psibetaprime}). Therefore, Beta-prime kernels are also $\mathcal{C}_c$-universal. Finally, $\mathcal{C}_0$-universality of Beta-prime kernels can be seen from Corollary \ref{c0universality} and (\ref{eq:betaprime}), since $k(x,y) \rightarrow 0$ when $x$ is fixed and $\Vert y \Vert \rightarrow \infty$ (for any matrix norm). 

As a technical aside, the spaces of real, complex, and quaternion matrices are not exactly symmetric spaces of non-compact type. However, each is a direct product of a symmetric space of non-compact type with a one-dimensional Euclidean space. This issue has no impact on the applicability of the results from Section \ref{main}. 

\section{Proofs}\label{proofs}

\begin{proof} [of Lemma \ref{kernelprops}]
By assumption, $\psi(\lambda,b)$ is a real number. Taking complex conjugates on both sides of (\ref{kernelnocomp}),
$$
k^*(x,y) = \int_\A \psi(\lambda, b) \, e^*(x,\lambda,b) \, e(y,\lambda,b) \,  d\nu(\lambda,b)\\
$$
However, this is just $k(y,x)$ by virtue of the same (\ref{kernelnocomp}). Thus, $k$ is clearly Hermitian.

The fact that $k$ is positive-definite follows from the further assumption that $\psi(\lambda,b) \geq 0$. Let~$(c_i;i=1,\ldots,p)$ be any complex numbers and $(x_i;i = 1,\ldots,p)$ be points in $X$. From (\ref{kernelnocomp}),
\begin{align*}
\sum^p_{i,j=1} c^{\phantom *}_i\hspace{0.02cm}c_j^*\hspace{0.03cm} k(x_i,x_j) & =
\int_\A \psi(\lambda, b) \left(\sum^p_{i,j=1} c^{\phantom *}_i\hspace{0.02cm}c_j^*\hspace{0.03cm} e(x_i,\lambda,b) \, e^*(x_j,\lambda,b)\right) \,  d\nu(\lambda,b) \\[0.1cm]
& =
\int_\A \psi(\lambda, b) \left| \sum^p_{i} c_i\hspace{0.03cm} e(x_i,\lambda,b)\right|^2 \,  d\nu(\lambda,b)
\end{align*}
which is clearly always a positive real number. Thus, $k$ is positive-definite. 

Finally, assume that $\psi(\lambda,b) = \psi(\lambda)$ is independent of $b \in B$. Recall the Plancherel measure $d\nu(\lambda,b) = |c(\lambda)|^{-2}d\lambda db$. In (\ref{kernelnocomp}), one may then integrate first over $b$ and then over $\lambda$, to obtain
\begin{equation} \label{eq:invarianceproof1}
k(x,y) = \int_{\mathfrak{a}^*_+} \psi(\lambda)\left(\,\int_B e(x,\lambda,b) \, e^* (y,\lambda,b) \hspace{0.03cm} db\right) |c(\lambda)|^{-2}\hspace{0.03cm} d\lambda 
\end{equation}
To show that $k$ is $G$-invariant, the crucial step is then to use the identity~\cite{helgason2022groups} (Page 418)
\begin{equation} \label{eq:invarianceproof2}
\int_B e(x,\lambda,b) \, e^* (y,\lambda,b) \hspace{0.03cm} db = \phi_\lambda(g^{-1}_1g^{\phantom{-1}}_2\!\!\!\!\cdot o)
\end{equation}
where $g_1,g_2 \in G$ are such that $x = g_1 \cdot o$ and $y = g_2 \cdot o$, and where $\phi_\lambda$ is the spherical function
\begin{equation} \label{eq:proofsphericalfunc}
\phi_\lambda(x) = \int_B\,e^*(x,\lambda,b)\,db
\end{equation}
From (\ref{eq:invarianceproof1}) and (\ref{eq:invarianceproof2}), 
$$
k(x,y) = \int_{\mathfrak{a}^*_+} \psi(\lambda)\hspace{0.03cm}\phi_\lambda(g^{-1}_1g^{\phantom{-1}}_2\!\!\!\!\cdot o)\hspace{0.03cm} |c(\lambda)|^{-2}\hspace{0.03cm} d\lambda 
$$
The fact that $k$ is $G$-invariant is now clear. Since $k(x,y)$ only depends on $g^{-1}_1g^{\phantom{-1}}_2\!\!$ and this remains unchanged if $x$ and $y$ are replaced by $g\dot x$ and $g \cdot y$ for any $g \in G$. 
\end{proof}

\begin{proof} [of Proposition \ref{l2godement}]
Let $k:X \times X \rightarrow \mathbb{C}$ be a square-integrable $G$-invariant positive-definite kernel. Consider the function $f:X \rightarrow \mathbb{C}$, where $f(x) = k(o,x)$. According to the $L^2$-Godement theorem \cite{dacosta2024invariantkernelsriemanniansymmetric},
$f$ may always be expressed 
\begin{equation} \label{eq:proof_L2G}
f(x) = \int_{\mathfrak{a}_+} \psi(\lambda)\hspace{0.03cm}\phi_\lambda(x)
\hspace{0.03cm}|c(\lambda)|^{-2}\hspace{0.03cm} d\lambda
\end{equation}
where $\psi:\mathfrak{a}^*_+ \rightarrow \mathbb{C}$ is positive-valued and belongs to $L^2(\mathfrak{a}^*_+,|c(\lambda)|^{-2}d\lambda)$, and where $\phi_\lambda$ is the spherical function (\ref{eq:proofsphericalfunc}). 
Now, because $k$ is $G$-invariant, it is clear $k(x,y) = f(g^{-1}_1g^{\phantom{-1}}_2\!\!\!\!\cdot o)$ for $g_1,g_2 \in G$ such that $x = g_1 \cdot o$ and $y = g_2 \cdot o$. Then, it follows from (\ref{eq:proof_L2G}),
$$
k(x,y) = \int_{\mathfrak{a}^*_+} \psi(\lambda)\hspace{0.03cm}\phi_\lambda(g^{-1}_1g^{\phantom{-1}}_2\!\!\!\!\cdot o)\hspace{0.03cm} |c(\lambda)|^{-2}\hspace{0.03cm} d\lambda 
$$
Using the identity (\ref{eq:invarianceproof2}) from the proof of Lemma \ref{kernelprops}, this last expression becomes
\begin{align*}
    k(x,y) & = \int_{\mathfrak{a}^*_+}\int_B \psi(\lambda)\hspace{0.03cm} e(x,\lambda,b) \, e^* (y,\lambda,b) \hspace{0.03cm}|c(\lambda)|^{-2}\hspace{0.03cm} d\lambda db \\[0.1cm]
           & = \int_{\mathcal{A}}\psi(\lambda)\hspace{0.03cm} e(x,\lambda,b) \, e^* (y,\lambda,b) \hspace{0.03cm} d\nu(\lambda,b) 
\end{align*}
just as in (\ref{kernelnocomp}). 
\end{proof}

\begin{proof} [of Lemma \ref{lemmarkhs}]
    The fact that $\mathcal{H}_k$ is a Hilbert space will follow by showing that $T:\mathcal{H}_k \rightarrow L^2(X)$ is an isomorphism of vector spaces. First, $T$ is linear and injective, since $\F$ is linear and injective, and since $\eta$ is $\nu$-a.e. non-zero. 

    Surjectivity of $T$ hinges on the fact that $\eta$ is bounded. For $h \in L^2(X)$, since $\F[h] \in L^2(\A)$, boundedness of $\eta$ implies $\eta\F[h] \in L^2(\A)$. Therefore, $f = \F^{-1}[\eta\F[h]]$ is well-defined. Moreover, for this $f$, $Tf = h$ so $h$ is in the image of $T$.

    Now, the scalar product on $\mathcal{H}_k$ is the pullback through $T$ of the scalar product on $L^2(X)$ (this works since $T$ is an isomorphism). Since $L^2(X)$ is a Hilbert space, $\mathcal{H}_k$ is a Hilbert space.    

    It remains to show $k$ is a reproducing kernel for $\mathcal{H}_k\hspace{0.03cm}$. For any $x \in X$, consider the function $k_x(y) = k(x,y)$. From (\ref{kernelnocomp}), 
    $$
    \F[k_x](\lambda,b) = \psi(\lambda, b)\,e(x,\lambda,b)
    $$
    which holds true since the right-hand side is in $L^2(\A)$, because  
    $e(x,\lambda,b)$ is bounded on $\A$. Then, 
    \begin{equation} \label{eq:kernelHFT}
    \frac{\F[k_x](\lambda,b)}{\eta(\lambda,b)} = \eta(\lambda, b)\,e(x,\lambda,b)
    \end{equation}
    Since $\eta \in L^2(\A)$, this shows that $\F[k_x]/\eta \in L^2(\A)$, which means that $k_x \in \mathcal{H}_k$ (by (\ref{rkhsnc})). Moreover, $k$ is indeed reproducing. For any $f\in \mathcal{H}_k$ and $x\in X$,
\begin{equation*}
\begin{split}
    \langle f,k_x\rangle_k:&= \langle Tf,Tk_x\rangle_{L^2(X)}\\[0.1cm]
    &= \langle \F[f]/\eta,\F[k_x]/\eta \rangle_{L^2(\mathcal{A})} \\[0.1cm]
    &= \int_\A  \, \frac{\F[f](\lambda,b)}{\eta(\lambda ,b)} \eta (\lambda, b)\, e^*(x,\lambda,b) \, d\nu(\lambda ,b) \\[0.1cm]
    &=\int_\A \F[f](\lambda,b)\,   e^*(x,\lambda,b) \, d\nu(\lambda ,b) = f(x) \\[0.1cm]
\end{split}
\end{equation*}
Here, the second equality follows from the isometry property of the Helgason-Fourier transform, the third equality by using (\ref{eq:kernelHFT}), and the very last one by the inversion formula for the Helgason-Fourier transform~\cite{helgason1970duality}.
\end{proof}

\begin{proof}[of Theorem \ref{ccuniversality}]
Note first that $\mathcal{H}_k$ is contained in $\mathcal{C}_b(X)$. Indeed, by the assumption that $\psi(\lambda) =\mathcal{O} (\langle \lambda \rangle^{-2s})$, any $f \in \mathcal{H}_k$ satisfies
\begin{equation*}
    \int_\A \langle\, \lambda \rangle^{2s} \, |\F [f](\lambda,b)|^2\, d\nu(\lambda,b) \leq c \,\int_\A \left(\psi(\lambda)\right)^{-1}|\F [f](\lambda,b)|^2\,d\nu(\lambda,b) <\infty
\end{equation*}
where $c$ is a constant which does not depend on the function $f$, and where the last inequality follows from (\ref{rkhsnc}). This means that $\langle \lambda\rangle^s\F[f] \in L^2(\A)$. Then, according to (\ref{eq:sobol_spectral}), $f \in H^s(X)$. However, since $s > n/2$, Theorem \ref{SET} implies $f \in \mathcal{C}_b(X)$. 

To show that $\mathcal{H}_k$ is dense in $\mathcal{C}(X)$ with respect to the topology of local uniform convergence, let $g$ be any continuous function, $g \in \mathcal{C}(X)$. The aim is to show, for any compact $C \subset X$ and $\epsilon > 0$, that there exists $f \in \mathcal{H}_k$ with $\| g-f\|_C < \epsilon$. Here, $\|\cdot\|_C$ denotes the 
seminorm $\| g-f\|_C = \sup_{x \in C}|g(x)-f(x)|$.

In a neighborhood of $C$, one may always uniformly approximate $g$ by a compactly-supported smooth function $h$. One way of obtaining this $h$ is to first multiply $g$ by a suitable bump function equal to $1$ on $C$, and then to smooth it by convolution with some smooth function. 

In this way, assume that $\|g-h\|_C < \epsilon/2$ where $h$ is compactly-supported and smooth. In~particular, $h \in H^s(X)$~\cite{hebey1996sobolev}. The next step is to approximate $h$ in the Sobolev norm by an element $f$ of $\mathcal{H}_k\hspace{0.03cm}$. 

By (\ref{eq:sobol_spectral}), $\langle \lambda \rangle^s \F[h]$ is in $L^2(\A)$. Because $\langle \lambda \rangle > 0$, there exist a compactly-supported continuous $\tilde{f}:\A \rightarrow \mathbb{C}$,
$$
\int_\A \langle \lambda \rangle^s \|\F[h](\lambda,b)-\tilde{f}(\lambda,b)\|^2\,d\nu(\lambda,b) < \epsilon/2\kappa
$$
where $\kappa$ is a constant to be defined below. If $f = \F^{-1}|\tilde{f}]$, this implies $\| h - f\|_{H^s(X)} < \epsilon/2\kappa$~\cite{pesenson2008discrete}. It~is also clear (by (\ref{rkhsnc})) that $f \in \mathcal{H}_k\hspace{0.03cm}$.  Finally, since $s > n/2$, the embedding $H^s(X) \hookrightarrow C_b(X)$ is continuous, if $\kappa$ is chosen to be its operator norm, then $\| h - f\|_C < \epsilon /2$. In conclusion, $\|g-f\|_C \leq \|g-h\|_C + \|h-f\|_C < \epsilon$, as required. 
\end{proof}

\begin{proof} [of corollary \ref{c0universality}]
The RKHS $\mathcal{H}_k$ is contained in $\mathcal{C}_0(X)$. This is because $k(x,\cdot) \in \mathcal{C}_0(X)$ for all $x \in X$~\cite{sriperumbudur2011universality}. For any $\epsilon > 0$, if $g \in \mathcal{C}_0(X)$, there exists $h \in \mathcal{C}_c(X)$, with $\|g-h\|_\infty < \epsilon/2$ ($\|\cdot\|_\infty$ the sumpremum norm). One may always assume $h$ is smooth, and therefore $h \in H^s(X)$.
Just as in the previous proof, there exists $f \in \mathcal{H}_k$ such that $\|h-f\|_\infty < \epsilon/2$. Then, $\|g-f\|_\infty < \epsilon$. Since $\epsilon$ and $g$ are arbitrary, this shows $\mathcal{H}_k$ is dense in $\mathcal{C}_{0}(X)$ with respect to the topology of uniform convergence. 
\end{proof}

\bibliographystyle{splncs04}
\bibliography{refs}
%

\end{document}